%% file: main.tex
\pdfoutput=1

\documentclass[11pt]{article}

\usepackage[]{acl}

\usepackage{times}
\usepackage{latexsym}

\usepackage[T1]{fontenc}

\usepackage[utf8]{inputenc}

\usepackage{microtype}
\usepackage{url}
\usepackage{makecell}
\usepackage{multirow}
\usepackage{graphicx}
\usepackage{caption}
\usepackage{subcaption}
\usepackage{amsmath}
\usepackage{bbm}
\usepackage{booktabs}
\usepackage{algorithmic}
\usepackage[ruled,vlined,linesnumbered]{algorithm2e}
\usepackage{microtype}
\usepackage{xcolor}
\usepackage{enumitem}

\title{Identifying and Mitigating Spurious Correlations\\ for Improving Robustness in NLP Models}

\author{
        Tianlu Wang \\ University of Virginia \\ tianlu@virginia.edu \And
        Rohit Sridhar\\ Georgia Institute of Technology \\ rohitsridhar@gatech.edu 
        \AND
        Diyi Yang \\ Georgia Institute of Technology \\ dyang888@gatech.edu \And 
        Xuezhi Wang \\ Google Research \\ xuezhiw@google.com}

\begin{document}
\maketitle
\begin{abstract}
Recently, NLP models have achieved remarkable progress across a variety of tasks; however, they have also been criticized for being not robust. Many robustness problems can be attributed to models exploiting \textit{spurious correlations}, or \textit{shortcuts} between the training data and the task labels.
Most existing work identifies a limited set of task-specific shortcuts via human priors or error analyses, which requires extensive expertise and efforts.
In this paper, we aim to automatically identify such spurious correlations in NLP models at scale. We first leverage existing interpretability methods to extract tokens that significantly affect model's decision process from the input text. We then distinguish ``genuine'' tokens and ``spurious'' tokens by analyzing model predictions across multiple corpora and further verify them through knowledge-aware perturbations. We show that our proposed method can effectively and efficiently identify a scalable set of ``shortcuts'',  and mitigating these leads to more robust models in multiple applications. 
\end{abstract}

\section{Introduction}
\label{sec:introduction}
\input{sections/introduction}

\section{Related Work}
\label{sec:related_work}
\input{sections/related_work.tex}

\section{Framework for Identifying Shortcuts}
\label{sec:method}
\input{sections/method.tex}

\section{Experiments}
\label{sec:experiment}
\input{sections/experiments.tex}

\section{Conclusion}
\input{sections/conclusion.tex}

\bibliography{anthology,ref}
\bibliographystyle{acl_natbib}

\appendix
\section{Appendix}
\label{sec:appendix}
\begin{figure}[ht]
        \centering
        \includegraphics[height=2.3in]{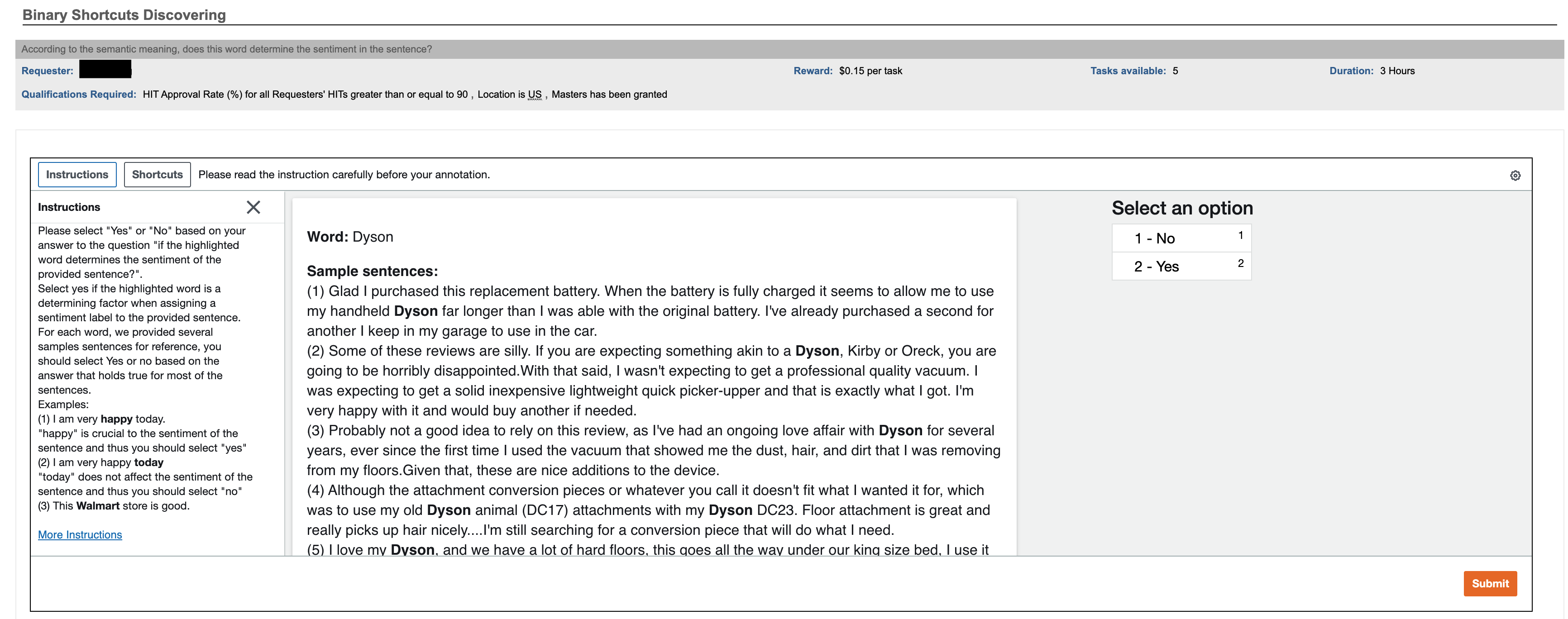}
    \caption{Amazon mechanical Turk interface.}
    \label{fig:amazon_turk}
\end{figure}
\end{document}

%% file: sections/introduction.tex
\begin{figure}[t]
        \centering
        \includegraphics[height=0.94in]{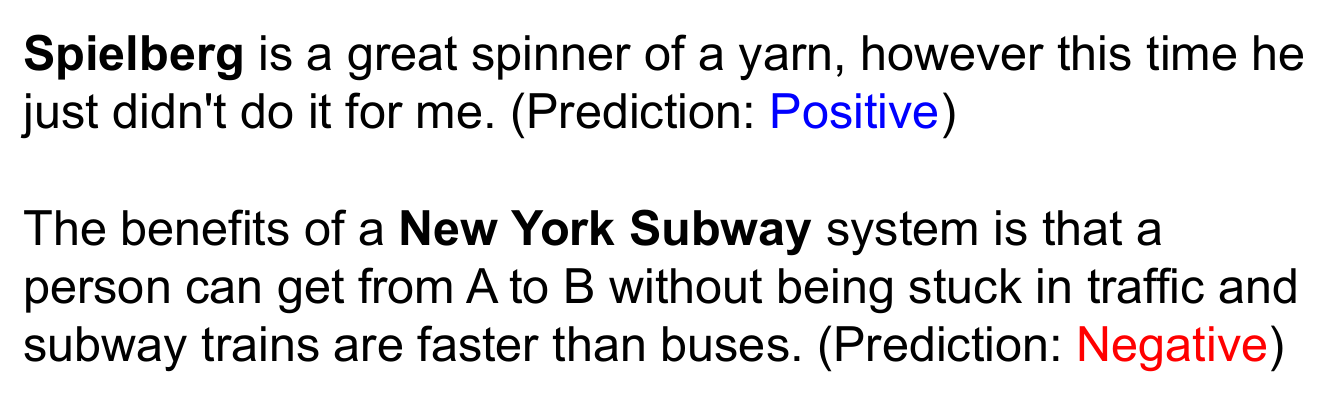}
\vspace{-0.25in}
    \caption{Examples of spurious correlations in sentiment classification task. A sentiment classification model takes \textit{Spielberg} and \textit{New York Subway} as shortcuts and makes wrong predictions.}
    \label{fig:teaser}
\vspace{-0.1in}
\end{figure}

Despite great progress has been made over improved accuracy, deep learning models are known to be brittle to out-of-domain data \cite{ood_robustness, wang-etal-2019-adversarial-domain}, adversarial attacks \cite{mccoy-etal-2019-right, jia-liang-2017-adversarial, textfooler}, partly due to sometimes the models have exploited \textit{spurious correlations} in the existing training data \cite{tu20tacl, sagawa2020investigation}.
In Figure~\ref{fig:teaser}, we show an example of a sentiment classification model making spurious correlations over the phrases ``\emph{Spielberg}'' and ``\emph{New York Subway}'' due to their high co-occurrences with positive and negative labels respectively in the training data.

\begin{table*}[h]
\begin{center}
\small
\setlength\tabcolsep{2.5pt} 
\begin{tabular}{c|c|c}
\toprule
 & Objective & Approach for shortcut identification  \\
\midrule
\citet{he2019unlearn} & Robustness against \textit{known} shortcuts & Pre-defined\\
\citet{clark2019don} & Robustness against \textit{known} shortcuts &  Pre-defined \\
\citet{clark2020learning} & Robustness against \textit{unknown} shortcuts & A low-capacity model to specifically learn shortcuts\\
\citet{wang-culotta-2020-identifying} & Identify \textit{unknown} shortcuts for robustness & A classifier over human annotated examples \\
This paper & Identify \textit{unknown} shortcuts for robustness & Automatic identification with interpretability methods \\
\hline
\end{tabular}
\end{center}
\vspace{-0.15in}
\caption{Comparison of our work and other related literature.
}
\vspace{-0.15in}
\label{tab:comparison}
\end{table*}

Most existing work quantifies spurious correlations in NLP models via a set of pre-defined patterns based on human priors and error analyses over the models, e.g., syntactic heuristics for Natural Language Inference \cite{mccoy-etal-2019-right}, synonym substitutions \cite{nl_adv_ucla}, or adding adversarial sentences for QA \cite{jia-liang-2017-adversarial}. More recent work on testing models' behaviour using CheckList \cite{ribeiro-etal-2020-beyond} also used a pre-defined series of test types, e.g., adding negation, temporal change, and switching locations/person names.
However, for safe deployment of NLP models in the real world, in addition to pre-defining a small or limited set of patterns which the model could be vulnerable to, it is also important to proactively \textit{discover} and \textit{identify} models' unrobust regions automatically and comprehensively. 

In this work, we introduce a framework to automatically identify spurious correlations  exploited by the model, sometimes also denoted as ``shortcuts'' in prior work \cite{Geirhos_2020, minderer2020automatic}\footnote{Throughout the paper we use spurious correlations and shortcuts interchangeably.}, at a large scale.
Our proposed framework differs from existing literature with a focus more on automatic shortcut identification, instead of pre-defining a limited set of shortcuts or learning from human annotations (Table~\ref{tab:comparison}).
Our framework works as follows: given a task and a trained model, we first utilize interpretability methods, e.g., attention scores \cite{clark-etal-2019-bert, kovaleva-etal-2019-revealing} and integrated gradient~\cite{pig} which are commonly used for interpreting model's decisions, to automatically extract tokens that the model deems as important for task label prediction. We then introduce two extra steps to further categorize the extracted tokens to be ``genuine'' or ``spurious''. We utilize a \textit{cross-dataset} analysis  to identify tokens that are more likely to be ``shortcut''. The intuition is that if we have data from multiple domains for the same task, then ``genuine'' tokens are more likely to remain useful to labels across domains, while ``spurious'' tokens would be less useful.
Our last step further applies a \textit{knowledge-aware perturbation} to check how stable the model's prediction is by perturbing the extracted tokens to their semantically similar neighbors.
The intuition is that a model's prediction is more likely to change when a ``spurious'' token is replaced by its semantically similar variations.
To mitigate these identified ``shortcuts'', we propose a simple yet effective \textit{targeted} mitigation approach to prevent the model from using those ``shortcuts'' and show that the resulting model can be more robust. Our code and data have been made publicly.\footnote{https://github.com/tianlu-wang/Identifying-and-Mitigating-Spurious-Correlations-for-Improving-Robustness-in-NLP-Models} Our contributions are as follows:

\begin{itemize}[leftmargin=5mm]
\setlength\itemsep{0.0in}
\vspace{-0.1in}
    \item We introduce a framework to automatically identify shortcuts in NLP models at scale. It first extracts important tokens using interpretability methods, then we propose \textit{cross-dataset analysis} and \textit{knowledge-aware perturbation} to distinguish spurious correlations from genuine ones.
    \item We perform experiments over several benchmark datasets and NLP tasks including sentiment classification and occupation classification, and show that our framework is able to identify more subtle and diverse spurious correlations. We present results showing the identified shortcuts can be utilized to improve robustness in multiple applications, including better accuracy over challenging datasets, better adaptation across multiple domains, and better fairness implications over certain tasks. 
\end{itemize}

%% file: sections/related_work.tex
\paragraph{Interpretability} 
There has been a lot of work on better interpreting models' decision process, e.g., understanding BERT \cite{clark-etal-2019-bert, kovaleva-etal-2019-revealing} and attention in transformers \cite{hao2020self-attention}, or through text generation models \cite{DBLP:journals/corr/abs-2004-14546}.
In this paper we utilize the attention scores as a generic way to understand what features a model relies on for making its predictions. Other common model interpretation techniques \cite{pig, LIME}, or more recent work on hierarchical attentions \cite{chen-etal-2020-generating} and contrastive explanations \cite{jacovi2021contrastive}, can be used as well.
In \citet{pruthi-etal-2020-learning}, the authors found that attention scores can be manipulated to deceive human decision makers. The reliability of existing interpretation methods is a research topic by itself, and extra care needs to be taken when using attention for auditing models on fairness and accountability \cite{fairwashing}.

\begin{figure*}[t]
        \centering
        \includegraphics[height=2.1in]{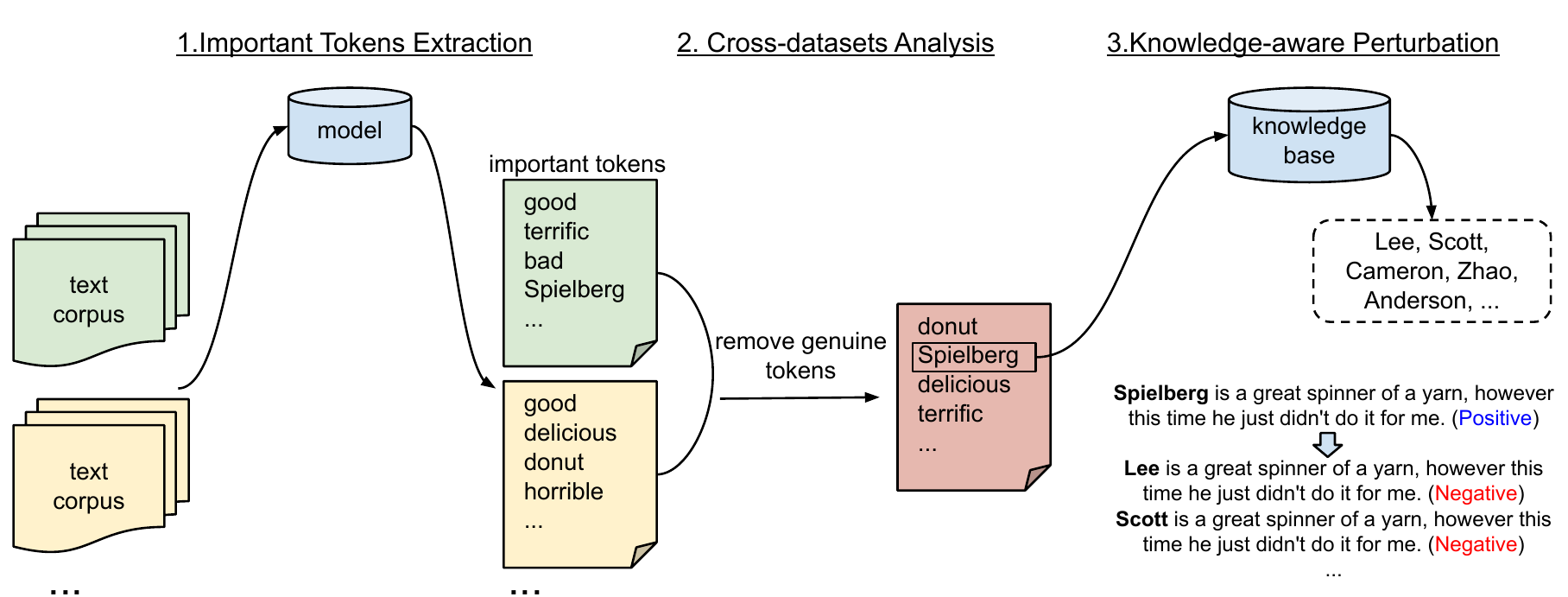}
    \caption{Our proposed pipeline to identify spurious correlations at scale. In the first step, we extract important tokens from input text. In the second step, we analyze extracted tokens from various datasets to identify likely ``spurious'' tokens. Finally, we further validate the output from the second step through knowledge-aware perturbation.}
    \label{fig:discovering}
\vspace{-0.1in}
\end{figure*}

\vspace{-0.1in}
\paragraph{Robustness and Bias} An increasing body of work has been conducted on understanding robustness in deep neural networks, particularly, how models sometimes might exploit spurious correlations \cite{tu20tacl, sagawa2020investigation} and take shortcuts \cite{Geirhos_2020}, leading to vulnerability in generalization to out-of-distribution data or adversarial examples in various NLP tasks such as NLI \cite{mccoy-etal-2019-right}, Question-Answering \cite{jia-liang-2017-adversarial}, and Neural Machine Translation \cite{niu-etal-2020-evaluating}.
Different from most existing work that defines types of spurious correlations or shortcut patterns beforehand \cite{ribeiro-etal-2020-beyond, mccoy-etal-2019-right, jia-liang-2017-adversarial}, which is often limited and requires expert knowledge, in this work we focus on automatically identifying models' unrobust regions at scale.
Another line of work aims at identifying shortcuts in models \cite{wang-culotta-2020-identifying} by training classifiers to better distinguish ``spurious'' correlations from ``genuine'' ones from human annotated examples. In contrast, we propose a cross-dataset approach and a knowledge-aware perturbation approach to automate this identification process with less human intervention in-between.

\vspace{-0.1in}
\paragraph{Mitigation} Multiple approaches have been proposed to mitigate shortcut learning and data biases \cite{clark2020learning, pmlr-v119-bras20a, zhou-bansal-2020-towards, minderer2020automatic}, through data augmentation \cite{textfooler, nl_adv_ucla}, domain adaptation \cite{blitzer-etal-2006-domain, blitzer-etal-2007-biographies}, and multi-task learning \cite{tu20tacl}. 
\citet{DBLP:journals/corr/abs-2103-06922} proposes to mitigate shortcuts by suppressing model's prediction on examples with a large shortcut degree.
Recent study has also shown removing spurious correlations can sometimes hurt model's accuracy \cite{10.1145/3442188.3445883}.
Orthogonal to existing works, we propose to first identify unrobust correlations in an NLP model and then propose a \textit{targeted} mitigation to encourage the model to rely less on those unrobust correlations.

%% file: sections/method.tex
In this section, we introduce our framework to identify spurious correlations in NLP models. 
Our overall framework consists of first identifying tokens important for models' decision process, followed by a cross-dataset analysis and a knowledge-aware perturbation step to identify spurious correlations. 

\subsection{Identify Tokens Key to Model's Decision}
The first step of the framework aims to identify the top-$K$ most important tokens that affect model's decision making process. 
We look at the importance at the token-level.\footnote{In this paper, we mostly focus on unigrams. Our method can also be easily extended to multi-gram, text span or other type of features by summing the attention scores over spans. For a vocabulary of wordpieces as used in BERT, we concatenate wordpieces with a prefix of ``\#\#'' to form unigrams and sum the attention scores.}
In general, depending on how the tokens are being used in model's decision process, they can be roughly divided into three categories: ``genuine'', ``spurious'', and others (e.g., tokens that are not useful for a model's prediction). 
\textit{Genuine} tokens are tokens that causally affect a task's label \cite{pmlr-v119-srivastava20a, zhao_spurious_2020}, and thus the correlations between those tokens and the labels are what we expect the model to capture and to more heavily rely on. 
On the other hand, \textit{spurious} tokens, or \textit{shortcuts} as commonly denoted in prior work \cite{Geirhos_2020, minderer2020automatic}, are features that correlate with task labels but are not genuine, and thus might fail to transfer to challenging test conditions \cite{Geirhos_2020} or out-of-distribution data; spurious tokens do not causally affect task labels \cite{pmlr-v119-srivastava20a, zhao_spurious_2020}.

In this step, we will extract both genuine tokens and shortcut tokens because they are both likely to affect a model's prediction.
We rely on interpretability techniques to collect information on whether a certain input token is important to model's decision making. In this paper, we use the attention score in BERT-based models as an explanation of model predictions \cite{clark-etal-2019-bert, kovaleva-etal-2019-revealing}, due to its simplicity and fast computation. Recent work~\cite{hiddencut} also reveals that attention scores outperform other explanation techniques in regularizing redundant information. 
Other techniques \cite{LIME, pig, chen-etal-2020-generating, jacovi2021contrastive} can also be used in this step. 
As an example, given a sentence ``\emph{Spielberg is a good director.}'', assuming ``\emph{good}'' is a genuine token and ``\emph{Spielberg}'' is a shortcut token, we expect that in a BERT-based sentiment classification model, the attention scores for ``\emph{good}'' and ``\emph{Spielberg}'' are higher and thus will be extracted as important tokens. On the other hand, for ``\emph{is}'', ``\emph{a}'' and ``\emph{director}'' the attention scores would be lower as they are relatively less useful to the model decision.

We now describe this step using sentiment classification task as an example (more details can be found in Algorithm~\ref{alg:step1}). Let $f$ be a well trained sentiment classification model. Given a corpus $\mathcal{D}$, for each input sentence $s_i$, $i=1, \ldots, n$ for a total of $n$ sentences in the corpus, we apply $f$ on it to obtain the output probability $p^{pos}_i$ and $p^{neg}_i$ for positive and negative label respectively. 
We then extract attention scores $\{a^1_{i}, a^2_{i}, \dots, a^m_{i}\}$ for tokens $\{t^1_{i}, t^2_{i}, \dots, t^m_{i}\}$ in sentence $s_i$, where $m$ is the length of the sentence. In BERT-based classification models, the embedding of [CLS] token in the final layer is fed to a classification layer. We thus extract the attention scores of each token $t$ used for computing the embedding of the [CLS] token and average them across different heads. 
If $p^{pos}_i > p^{neg}_i$, we obtain the updated attention score $\Tilde{a}^j_i = a^j_i * p^{pos}_i$, otherwise $\Tilde{a}^j_i = -a^j_i * p^{neg}_i$.
For each token $t$ in the vocabulary $\mathcal{V}$, we compute the average attention score: $\Bar{a}_t = \frac{1}{mn} \cdot \Sigma_{i=1}^n\Sigma_{j=1}^{m}[\Tilde{a}^j_i \cdot \mathbbm{1}(t^j_i=t)]$,
where we aggregate the attention scores $\Tilde{a}^j_i$ for token $t$, across all $n$ sentences in the corpus. 
We then normalize the attention scores across the vocabulary to obtain the importance score for each token $t$: $I_t = \Bar{a}_t / \Sigma_{t\in \mathcal{V}}\Bar{a}_{t}$. This can lead to very small $I_t$ for certain tokens, thus we take the log of all importance scores to avoid underflow, $I'_t = \log(I_t)$.

So far, we have computed the importance score for each token. However, we observe that some tokens appearing only very a few times could accidentally have very high importance scores. Thus, we propose to penalize the tokens with low frequencies:
$\hat{I}_t = I'_t - \lambda/\log(1+c_t),$
where $c_t$ is the frequency of token $t$ and $\lambda$ is a temperature parameter to adjust the degree that we want to penalize over the frequency.

\begin{algorithm}[t]
\SetAlgoLined
\SetKwInOut{Input}{Input}\SetKwInOut{Output}{Output}
\Input{Sentiment classification model: $f$\\
Text corpus: $\mathcal{D}$}
// Obtain attention scores for tokens in each input sentence $s_i\in\mathcal{D}$:\\
\For{$i = 1$ to n}{
$p_i^{pos}$, $p_i^{neg}$, $\{a_i^1, a_i^2, ..., a_i^m\}$  = $f(s_i)$\;
\For{$j = 1$ to m}{
if $p^{pos}_i > p^{neg}_i$:
    $\Tilde{a}^j_i = a^j_i \cdot p^{pos}_i$\;
else:
    $\Tilde{a}^j_i = -a^j_i \cdot p^{neg}_i$\;
}
}
// Use $\{\Tilde{a}_i^j\}$ to compute an importance score for each token $t$ in the vocabulary $\mathcal{V}$:\\
$Importance = dict()$\\
\For{$i = 1$ to n}{
\For{$j = 1$ to m}{
    $Importance[t_i^j]$.append($\Tilde{a}_i^j$)\;
    }
 }
// Normalize the importance score and penalize low-frequency tokens:\\
\For{$t$ in $\mathcal{V}$}{
    $\Bar{a}_t = average(Importance[t])$\;
    $I_t = \Bar{a}_t / \Sigma_{t\in \mathcal{V}}\Bar{a}_{t}$\;
    $I'_t = \log(I_t)$\;
    $\hat{I}_t = I'_t - \lambda / \log(1 + frequency[t])$\;
 }
\Output{A list of tokens sorted according to their importance scores:\\
\{$t_1, t_2, ..., t_{|\mathcal{V}|}$\}, \\
where $\hat{I}_{t_i} \geq \hat{I}_{t_2} \geq ... \geq \hat{I}_{t_{|\mathcal{V}|}}$
}
 \caption{Important Token Extraction.}
 \label{alg:step1}
\end{algorithm}

\begin{table*}[t]
\begin{center}
\small
\setlength\tabcolsep{2.5pt} 
\begin{tabular}{l}
\toprule
\textit{Shortcut token}: \textbf{bread}\\
\makecell[l]{\textit{Original:} I bought this in the hopes it would keep \textbf{bread} I made fresh. However, after a few times of usings the I found out \\that moister w still getting in bread would become stale or moldy ...(\textcolor{red}{Neg})}\\
\makecell[l]{\textit{Perturbed:} I bought this in the hopes it would keep \textbf{loaf} I made fresh. However, after a few times of usings the I found out \\that moister w still getting in bread would become stale or moldy ... (\textcolor{blue}{Pos})}\\
\midrule
\textit{Shortcut token}: \textbf{iPhone} \\
\makecell[l]{\textit{Original:} I lost my original TV remote, and found this one thinking it was the same one.  ... Now this one is merely a back \\up. Also, I have the Samsung remote app on my \textbf{iPhone}, which also works just as good as these remotes. (\textcolor{blue}{Pos})}\\
\makecell[l]{\textit{Perturbed:} I lost my original TV remote, and found this one thinking it was the same one.  ... Now this one is merely a back \\up. Also, I have the Samsung remote app on my \textbf{ipod}, which also works just as good as these remotes. (\textcolor{red}{Neg})}\\
\bottomrule
\end{tabular}
\end{center}
\vspace{-0.1in}
\caption{Examples of shortcut tokens with significant performance drop during knowledge-aware perturbation.}
\label{tab:qualitative}
\vspace{-0.1in}
\end{table*}

\subsection{Cross-Dataset Stability Analysis}
\label{sec:cross-dataset}
As mentioned before, the tokens that are important to a model's prediction could be either genuine or spurious, thus in this step, we want to categorize the extracted tokens into these two categories and maintain a list of tokens that are more likely to be ``spurious''. 

In many real-world NLP tasks, if we have access to datasets from different sources or domains, then we can perform a \textit{cross-dataset} analysis to more effectively identify ``spurious'' tokens.
The reasoning is that ``spurious'' tokens tend to be important for a model's decision making on one dataset but are less likely to transfer or generalize to other datasets, e.g. ``\emph{Spielberg}'' could be an important token for movie reviews but is not likely to be useful on other review datasets (e.g., for restaurants or hotels). On the other hand, genuine tokens are more likely to be important across multiple datasets, for example, tokens like ``\emph{good}'', ``\emph{bad}'', ``\emph{great}'', ``\emph{terrible}'' should remain useful across various sentiment classification datasets. Thus, in this step, we try to distinguish ``genuine'' tokens from ``spurious'' tokens from the top extracted important tokens after the first step. Our idea is to compare tokens' importance ranking and find the ones that have very different ranks across datasets.

To this end, we conduct a cross-dataset stability analysis. Specifically, we apply the same model $f$ on two datasets A and B, and obtain two importance ranking lists. Since importance scores may have different ranges on the two datasets, we normalize all importance scores to adjust the value to be in the range of $[0, 1]$:
$$\Tilde{I}^A_t = \frac{\hat{I}^A_t - \min(\{\hat{I}^A_t|t\in\mathcal{V}\})}{\max(\{\hat{I}^A_t|t\in\mathcal{V}\}) - \min(\{\hat{I}^A_t|t\in\mathcal{V}\})}$$
$$\Tilde{I}^B_t = \frac{\hat{I}^B_t - \min(\{\hat{I}^B_t|t\in\mathcal{V}\})}{\max(\{\hat{I}^B_t|t\in\mathcal{V}\}) - \min(\{\hat{I}^B_t|t\in\mathcal{V}\})}$$
where $\Tilde{I}^A_t$ and $\Tilde{I}^B_t$ are normalized importance scores on dataset A and B respectively. We then subtract $\Tilde{I}^B_t$ from $\Tilde{I}^A_t$ and re-rank all tokens according to their differences. Tokens with largest differences are the ones with high importance scores in dataset A but low importance scores in dataset B, thus they are more likely to be ``shortcut'' tokens in dataset A. Similarly, we can also extract tokens with largest differences from dataset B by subtract $\Tilde{I}^A_t$ from $\Tilde{I}^B_t$.

\subsection{Knowledge-aware Perturbation}
The cross-dataset analysis is an efficient way to remove important tokens that are ``genuine'' across multiple datasets, after which we can obtain a list with tokens that are more likely to be ``spurious''. However, on this list, domain-specific genuine tokens can still be ranked very high, e.g., ``\emph{ambitious}'' from a movie review dataset and ``\emph{delicious}'' from a restaurant review dataset. This is because domain-specific genuine tokens have similar characteristics as shortcuts, they are effective for a model's decision making on a certain dataset but could appear very rarely (and thus could be deemed as not important) on another dataset. 
Hence, in this section, we further propose a slightly more expensive and a more fine-grained approach to verify whether a token is indeed ``spurious'', through \textit{knowledge-aware perturbation}. 

For each potential shortcut token, we extract $N$ synonyms by leveraging the word embeddings curated for synonym extraction~\cite{counter-fitting}, plus WordNet~\cite{wordnet} and DBpedia~\cite{10.5555/1785162.1785216}. More specifically, for each top token $t$ in the list generated by the previous step, we first search counter-fitting word vectors to find synonyms with cosine similarity larger than a threshold\footnote{We set it as $0.5$ following the set up in~\cite{textfooler}.} $\tau$. Additionally we search in WordNet and DBpedia to obtain a maximum of $N$ synonyms for each token $t$. Then we extract a subset $S_t$ from $\mathcal{D}$, which consists of sentences containing $t$. We perturb all sentences in $S_t$ by replacing $t$ with its synonyms. The resulted perturbed set $S'_t$ is $N$ times of the original set $S_t$. We apply model $f$ on $S_t$ and $S'_t$ and obtain accuracy $acc_t$ and $acc'_t$. Since we only perturb $S_t$ with $t$'s synonyms, the semantic meaning of perturbed sentences should stay close to the original sentences. Thus, if $t$ is a genuine token, $acc'_t$ is expected to be close to $acc_t$. On the other hand, if $t$ is a shortcut, model prediction can be different even the semantic meaning of the sentence does not change a lot (see examples in Table~\ref{tab:qualitative}). Thus, we assume tokens with larger differences between $acc_t$ and $acc'_t$ are more likely to be shortcuts and tokens with smaller differences are more likely to be domain specific ``genuine'' words. From the potential shortcut token list computed in Sec~\ref{sec:cross-dataset}, we remove tokens with performance difference smaller than $\mathbbm{\delta}$ to further filter domain specific ``geniue'' tokens .

\begin{table*}[t]
\begin{center}
\small
\setlength\tabcolsep{2.5pt} 
\begin{tabular}{c|c}
\toprule
Dataset & Top \textbf{important} tokens extracted from each dataset \\
\midrule
SST-2 & terrific, impeccable, exhilarating, refreshingly, irresistible, heartfelt, thought-provoking, ...\\
Yelp & Awesome, Definitely, is, Excellent, Very, Great, Good, Best, attentive, worth, definitely, Highly, ...\\
Amazon Kitchen & utensils, thermometer, Cuisinart, Definitely, Pyrex, Bought, utensil, Arrived, Recommend, ...\\
\midrule
Dataset & Top \textbf{shortcuts} extracted from each dataset (verified by human annotators) \\
\midrule
SST-2 & recycled, seal, sitcom, longest, fallen, qualities, rises, impact, translate, emphasizes, ...\\
Yelp & ambiance, tastes, bartenders, patio, burgers, staff, watering, donuts, cannot, pancakes, regulars, ...\\
Amazon Kitchen & utensils, Cuisinart, Rachael, Pyrex, utensil, Breville, Zojirushi, Corelle, Oxo, dehumidifier, ...\\
\bottomrule
\end{tabular}
\end{center}
\vspace{-0.15in}
\caption{Top \textit{important} tokens and top \textit{shortcut} tokens identified by our proposed framework and further verified by human annotators. Many shortcuts reflect the characteristics of the datasets, e.g. ``captures'' from a movie review dataset, ``burgers'' from a restaurant review dataset and brand names from an Amazon kitchen review dataset. }
\label{tab:important_words}
\end{table*}

\begin{table*}[htb]
\begin{center}
\small
\begin{tabular}{c|c|cc|cc|cc}
\toprule
\multirow{2}{*}{Dataset} & \multirow{2}{*}{Method} & \multicolumn{2}{c|}{@10} &  \multicolumn{2}{c|}{@20} &  \multicolumn{2}{c}{@50} \\
  & & Prec. & Impor. & Prec. & Impor. & Prec. & Impor. \\
\midrule
\multirow{3}{*}{SST-2} &  1 & 0.00 & - & 0.05 & 0.97 & 0.02 & 0.96\\
&  2 & 0.10 & 0.95 & 0.05 & 0.94 & 0.04 & 0.93\\
&  3 & 0.40 & 0.90 & 0.35 & 0.87 & 0.32 & 0.85\\

\midrule
\multirow{3}{*}{Yelp} &  1 & 0.10 & 0.96 & 0.05 & 0.95 & 0.18 & 0.95\\
&  2 & 0.40 & 0.89 & 0.25 & 0.89 & 0.30 & 0.88\\
&  3 & 0.60 & 0.89 & 0.50 & 0.87 & 0.56 & 0.87\\
\midrule
\multirow{3}{*}{\makecell{Amazon Kitchen}} &  1 & 0.70 & 0.98 & 0.80 & 0.96 & 0.78 & 0.95\\
&  2 &1.00 & 0.97 & 1.00 & 0.95 & 1.00 & 0.95\\
&  3 &1.00 & 0.97 & 1.00 & 0.95 & 1.00 & 0.95\\
\bottomrule            
\end{tabular}
\end{center}
\vspace{-0.15in}
\caption{We report the precision as well as the averaged importance score $\Tilde{I}$ of identified ``shortcuts'' after each step based on our framework. The identified ``shortcut'' is a true shortcut or not is verified by 3 independent human annotators (Amazon Turkers). We can see that the precision increases after each step in our framework, demonstrating the utility of cross-dataset analysis (step 2) and   knowledge-aware perturbation (step 3). }
\label{tab:precision_exp}
\vspace{-0.1in}
\end{table*}

\subsection{Mitigation via Identified Shortcuts}
\label{sec:mitigate}
In this section, we describe how the identified shortcuts can be further utilized to improve robustness in NLP models.
More specifically, we propose \textit{targeted} approaches to mitigate the identified shortcuts including three variants: 
(1) a \textit{training-time} mitigation approach: we mask out the identified shortcuts during training time and re-train the model;
(2) an \textit{inference-time} mitigation approach: we mask out the identified shortcuts during inference time only, in this way we save the extra cost of re-training a model;
(3) we combine both approach (1) and (2).
In the experiment section, we will demonstrate the effect of each approach over a set of benchmark datasets. We found that by masking out shortcuts in datasets, models generalize better to challenging datasets, out-of-distribution data, and also become more fair.

%% file: sections/experiments.tex
\subsection{Tasks and Datasets}
\noindent \textbf{Task 1: Sentiment classification.} For the task of sentiment classification, we use several datasets in our experiments. To find shortcuts in Stanford Sentiment Treebank (SST-2) \cite{socher-etal-2013-recursive} dataset, we first train a model on SST-2 training set which consists of $67,349$ sentences. We then evaluate the model on SST-2 training set\footnote{We use training set of SST-2 because the test set has a very limited number of examples.} and Yelp~\cite{asghar2016yelp} test set and obtain attention scores. For cross-dataset analysis, we compare the important tokens extracted from SST-2 and Yelp.
Similarly, we train another model on $80,000$ amazon kitchen reviews \cite{amazon_data}, and apply it on the kitchen review dev set and the amazon electronics dev set, both having $10,000$ reviews. 

\noindent \textbf{Task 2: Occupation classification.} Following \citet{pruthi-etal-2020-learning}, we use the biographies \cite{De_Arteaga_2019} to predict whether the occupation is a surgeon or physician (non-surgeon). The training data consists of $17,629$ biographies and the dev set contains $2,519$ samples.

\noindent \textbf{Models.}
We use the attention scores over BERT \cite{devlin-etal-2019-bert} based classification models as they have achieved the state-of-art performance. Note that our proposed framework can also be easily extended to models with different architectures. BERT-based models have the advantage that we can directly use the attention scores as explanations of model decisions. For models with other architectures, we can use explanation techniques such as LIME~\cite{LIME} or Path Integrated Gradient approaches~\cite{pig} to provide explanations.

\noindent \textbf{Evaluation.}
Evaluating identified shortcuts in machine learning or deep leaning based models can be difficult.
We do not have ground-truth labels for the shortcuts identified through our framework, and whether a token is a shortcut or not can be subjective even with human annotators, and it can further depend on the context.
Faced with these challenges, we carefully designed a task and adopted Amazon Mechanical Turk for evaluation. We post the identified shortcuts after each step in our framework, along with several sample sentences containing the token, as additional context, to the human annotator. We ask the question ``does the word determine the sentiment in the sentence'' and ask the annotator to provide a ``yes''/``no'' answer\footnote{In the instruction, we further specify ``select yes'' if the highlighted word is a determining factor for the sentiment label, and we provide a few example sentences along with their shortcuts as references. The exact template is shown in the Appendix.} to the question based on the answer that holds true for the majority of the provided sentences (we also experimented with adding an option of ``unsure'' but found most annotators do not choose that option). Each identified shortcut is verified by 3 annotators.

\begin{table*}[t!]
\begin{center}
\small
\begin{tabular}{c|c|c|c|c|c|c}
\toprule
Methods &  \makecell{SST-2 $\rightarrow$\\ Kitchen} &
\makecell{SST-2 $\rightarrow$\\ Electronics}& \makecell{ Kitchen $\rightarrow$\\ SST-2} & \makecell{Kitchen $\rightarrow$\\ Electronics} & \makecell{Electronics $\rightarrow$\\ SST-2}&
\makecell{Electronics $\rightarrow$\\ Kitchen}\\
\midrule
No Mitigation &87.43 & \textbf{84.30} & 71.45 & 98.22 &73.05 & 98.79\\
Test RM &87.50 & 83.96 & 71.56 & 98.07 & 72.94 & 98.77\\
Train RM & 87.72 & 84.13 & \textbf{72.82} &\textbf{98.60} &  \textbf{74.08} & 98.79\\
Train \& Test RM & \textbf{87.76} & 83.74 & \textbf{72.82} & \textbf{98.62} & \textbf{74.08} & \textbf{98.80}\\
\bottomrule            
\end{tabular}
\end{center}
\vspace{-0.1in}
\caption{Domain generalization results on SST-2 and Amazon Kitchen/Electronics datasets. RM means shortcuts removed, Train/Test corresponds to shortcuts removal during training and test time, respectively.}
\label{tab:transfer}
\end{table*}

\subsection{Experimental Results}
We summarized the top important tokens after each step in our framework (Table~\ref{tab:important_words}). We also report the precision score (the percentage of tokens) out of the top $50$ tokens identified as true shortcuts by human annotators in Table~\ref{tab:precision_exp}.

Across all datasets, we see that the precision score increases after each step, which demonstrates that our proposed framework can consistently improve shortcut identification more precisely. Specifically, after the first step, the precision score of shortcuts is low\footnote{Some cells have ``-'' importance score due to no shortcut is identified by human annotators in the top-$K$ identified tokens.} because most of the top extracted tokens are important tokens only (thus many of them are genuine). After the second step (cross-dataset analysis) and the third step (knowledge-aware perturbation), we see a significant increase of the shortcuts among the top-$K$ extracted tokens. Table~\ref{tab:qualitative} shows examples of perturbing shortcut tokens leading to model predictions changes. 

\textbf{Agreement analysis over annotations.} Since this annotation task is non-trivial and sometimes subjective, we further compute the intraclass correlation score~\cite{bartko1966intraclass} for the Amazon Mechanical Turk annotations. Our collected annotations reaches an intraclass correlation score of $0.72$, showing a good agreement among annotators. Another agreement we analyze is showing annotators $5$ sample sentences compared to showing them all sentences, to avoid sample bias. We ask annotators to annotate a batch of $25$ tokens with all sentences containing the corresponding token shown to them. The agreement reaches $84.0\%$, indicating that showing $5$ sample sentences does not significantly affect annotator's decision on the target token. More details of Amazon Mechanical Turk interface can be found in the Appendix.

\begin{table}[t]
\begin{center}
\small
\begin{tabular}{c|c|c}
\toprule
Top $10$ extracted tokens & Precision        & Recall        \\
\midrule
\makecell{\colorbox{yellow!50}{ms.}, \colorbox{yellow!50}{mrs.}, \colorbox{yellow!50}{she}, \colorbox{yellow!50}{her}, \\ \colorbox{yellow!50}{he}, reviews, been, \\favorite, \colorbox{yellow!50}{his}, practices} & 0.60 & 0.50\\
\bottomrule            
\end{tabular}
\end{center}
\vspace{-0.1in}
\caption{Identified shortcuts (highlighted tokens are overlapped with the pre-specified impermissible tokens from \citet{pruthi-etal-2020-learning}) in occupation classification.}
\label{tab:occupation}
\end{table}
\subsection{A Case Study: Occupation Classification}
\citet{pruthi-etal-2020-learning} derived an occupation dataset to study the gender bias in NLP classification tasks. The task is framed as a binary classification task to distinguish between ``surgeons'' and ``physicians''. These two occupations are chosen because they share similar words in their biographies and a majority of surgeons are male. The dataset is further tuned -- downsample minority classes (female surgeons and male physicians) by a factor of ten to encourage the model to rely on gendered words to make predictions. \citet{pruthi-etal-2020-learning} also provides a pre-specified list of impermissible tokens~\footnote{he, she, her, his, him, himself, herself, mr, ms, mr., mrs., ms. We removed ``hers'' and ``mrs'' from the original list since they do not appear in dev data.} that a robust model should assign low attention scores to. We instead treat this list of tokens as shortcuts and analyze the efficacy of our proposed framework on identifying these tokens. These impermissible tokens can be regarded as shortcuts because they only reflect the gender of the person, thus by definition should not affect the decision of a occupation classification model.
Table~\ref{tab:occupation} presents the result on identifying the list of impermissible tokens. Among the top ten tokens selected by our method, $6$ of them are shortcuts. 
Furthermore, $9$ out of $12$ impermissible tokens are captured in the top $50$ tokens selected by our method. 
This further demonstrates that our method can effectively find shortcuts in this occupation classification task, in a more automated way compared to existing approaches that rely on pre-defined lists.

\begin{table}[t!]
\begin{center}
\small
\begin{tabular}{c|c|c|c}
\toprule
Dataset & Methods & C1 & C2\\
\midrule
\multirow{4}{*}{\makecell{Amazon \\ Kitchen}} & No Mitigation & 99.15 & 0.0\\
&  Test RM & 99.21 & 0.18\\
& Train RM & 99.15 & 0.24\\
& Train \& Test RM & 99.15 & 0.24\\
\bottomrule            
\end{tabular}
\end{center}
\vspace{-0.1in}
\caption{Accuracy on challenging datasets. C1: test subset that has shortcuts; C2: test subset that has shortcuts and are wrongly predicted by the original model.}
\label{tab:challenging}
\end{table}

\begin{table}[t!]
\begin{center}
\small
\begin{tabular}{c|cccc}
\toprule
 & Male & Female & $\Delta$ & Overall \\
\midrule
No Mitigation & 94.02 & 99.50 & 5.48 & 97.46 \\
Test RM & 92.28 & 96.36 & 4.08 & 94.84 \\
Train RM & 93.26 & 92.40 & 0.86 & 92.66 \\
Train \& Test RM & 94.46 & 99.06 & 4.60 & 97.34\\
\bottomrule            
\end{tabular}
\end{center}
\vspace{-0.1in}
\caption{Accuracy and performance gap of male and female groups in Occupation Classification task.}
\label{tab:bias}
\vspace{-0.1in}
\end{table}

\begin{table}[t]
\begin{center}
\small
\begin{tabular}{c|c|c|c|c}
\toprule
$\lambda$ & 4 & 6 & 8 & 10 \\
\midrule
4 & 1.00 & 0.78 & 0.62 & 0.56\\
6 & 0.78 & 1.00 & 0.84 & 0.76\\
8 & 0.62 & 0.84 & 1.00 & 0.92\\
10 & 0.56 & 0.76 & 0.92 & 1.00\\
\bottomrule            
\end{tabular}
\end{center}
\vspace{-0.1in}
\caption{Overlap of top $50$ tokens when changing $\lambda$. 
}
\label{tab:ablation}
\end{table}

\begin{table}[htb]
\begin{center}
\small
\begin{tabular}{c|c|c|c|c}
\toprule
\multirow{2}{*}{Dataset} & \multirow{2}{*}{Method} & \multicolumn{1}{c|}{@10} &  \multicolumn{1}{c|}{@20} &  \multicolumn{1}{c}{@50} \\
  & &  Prec. & Prec. & Prec. \\
\midrule
\multirow{2}{*}{SST-2} & Attention & 0.40 & 0.35 & 0.32 \\
& Integrated Gradient & 0.30 & 0.3 & 0.34 \\
\midrule
\multirow{2}{*}{Yelp} & Attention & 0.60 & 0.50 & 0.56 \\
& Integrated Gradient & 0.50 & 0.55 & 0.60 \\
\bottomrule            
\end{tabular}
\end{center}
\vspace{-0.1in}
\caption{Ablation study on using Integrated Gradient to extract important tokens.}
\label{tab:ablation_ig}
\vspace{-0.1in}
\end{table}

\subsection{Mitigating Shortcuts}
We also study mitigating shortcuts by masking out the identified shortcuts. Specifically, we use shortcut tokens identified by human annotators and mask them out in training set  and re-train the model (Train RM), during test time directly (Test RM), and both (Train \& Test RM) as described in Sec~\ref{sec:mitigate}. We evaluate these three approaches in multiple settings: 1) domain generalization; 2) challenging datasets; 3) gender bias. As shown in Table~\ref{tab:transfer}, masking out shortcuts, especially in training data, can improve model's generalization to out-of-distribution data.
Note in this setting, different from existing domain transfer work \cite{pan_transfer_learning_survey}, we do not assume access to labeled data in the target domain during training, instead we use our proposed approach to identify potential shortcuts that can generalize to unseen target domains.
As a result, we also observe model's performance improvement on challenging datasets (Table~\ref{tab:challenging}). Table~\ref{tab:bias} demonstrates that mitigating shortcuts helps to reduce the performance gap ($\Delta$) between male and female groups, resulting in a fairer model.
Note the original performance might degrade slightly due to models learning different but more robust feature representations, consistent with findings in existing work \cite{tsipras2018robustness}.

\paragraph{Ablation Study}
We conduct an ablation study of changing the hyper-parameter $\lambda$ in the first step of extracting important tokens. As shown in Table~\ref{tab:ablation}, our method is not very sensitive to the choice of $\lambda$. In Table~\ref{tab:ablation_ig}, we show that Attention scores and Integrated Gradient can both serve as a reasonable method for extracting important tokens in our first step, suggesting the flexibility of our framework.

%% file: sections/conclusion.tex
\label{sec:conclusion}
In this paper, we aim to improve NLP models' robustness via identifying spurious correlations automatically at scale, and encouraging the model to rely less on those identified shortcuts.
We perform experiments and human studies over several benchmark datasets and NLP tasks to show a scalable set of shortcuts can be efficiently identified through our framework.
Note that we use existing interpretability approaches as a proxy to better understand how a model reaches its prediction, but as pointed out by prior work, the interpretability methods might not be accurate enough to reflect how a model works (or sometimes they could even deceive human decision makers). We acknowledge this as a limitation, and urge future research to dig deeper and develop better automated methods with less human intervention or expert knowledge in improving models' robustness.

%% file: main.bbl
\begin{thebibliography}{46}
\expandafter\ifx\csname natexlab\endcsname\relax\def\natexlab#1{#1}\fi

\bibitem[{Alzantot et~al.(2018)Alzantot, Sharma, Elgohary, Ho, Srivastava, and
  Chang}]{nl_adv_ucla}
Moustafa Alzantot, Yash Sharma, Ahmed Elgohary, Bo-Jhang Ho, Mani Srivastava,
  and Kai-Wei Chang. 2018.
\newblock \href {https://doi.org/10.18653/v1/D18-1316} {Generating natural
  language adversarial examples}.
\newblock In \emph{Proceedings of the 2018 Conference on Empirical Methods in
  Natural Language Processing}, pages 2890--2896, Brussels, Belgium.
  Association for Computational Linguistics.

\bibitem[{Asghar(2016)}]{asghar2016yelp}
Nabiha Asghar. 2016.
\newblock \href {http://arxiv.org/abs/1605.05362} {Yelp dataset challenge:
  Review rating prediction}.

\bibitem[{Auer et~al.(2007)Auer, Bizer, Kobilarov, Lehmann, Cyganiak, and
  Ives}]{10.5555/1785162.1785216}
S\"{o}ren Auer, Christian Bizer, Georgi Kobilarov, Jens Lehmann, Richard
  Cyganiak, and Zachary Ives. 2007.
\newblock Dbpedia: A nucleus for a web of open data.
\newblock In \emph{Proceedings of the 6th International The Semantic Web and
  2nd Asian Conference on Asian Semantic Web Conference}, ISWC'07/ASWC'07, page
  722–735, Berlin, Heidelberg. Springer-Verlag.

\bibitem[{Aïvodji et~al.(2019)Aïvodji, Arai, Fortineau, Gambs, Hara, and
  Tapp}]{fairwashing}
Ulrich Aïvodji, Hiromi Arai, Olivier Fortineau, Sébastien Gambs, Satoshi
  Hara, and Alain Tapp. 2019.
\newblock Fairwashing: the risk of rationalization.
\newblock In \emph{Proceedings of the 34th International Conference on Machine
  Learning}.

\bibitem[{Bartko(1966)}]{bartko1966intraclass}
John~J Bartko. 1966.
\newblock The intraclass correlation coefficient as a measure of reliability.
\newblock \emph{Psychological reports}, 19(1):3--11.

\bibitem[{Blitzer et~al.(2007)Blitzer, Dredze, and
  Pereira}]{blitzer-etal-2007-biographies}
John Blitzer, Mark Dredze, and Fernando Pereira. 2007.
\newblock \href {https://www.aclweb.org/anthology/P07-1056} {Biographies,
  {B}ollywood, boom-boxes and blenders: Domain adaptation for sentiment
  classification}.
\newblock In \emph{Proceedings of the 45th Annual Meeting of the Association of
  Computational Linguistics}, pages 440--447, Prague, Czech Republic.
  Association for Computational Linguistics.

\bibitem[{Blitzer et~al.(2006)Blitzer, McDonald, and
  Pereira}]{blitzer-etal-2006-domain}
John Blitzer, Ryan McDonald, and Fernando Pereira. 2006.
\newblock \href {https://www.aclweb.org/anthology/W06-1615} {Domain adaptation
  with structural correspondence learning}.
\newblock In \emph{Proceedings of the 2006 Conference on Empirical Methods in
  Natural Language Processing}, pages 120--128, Sydney, Australia. Association
  for Computational Linguistics.

\bibitem[{Bras et~al.(2020)Bras, Swayamdipta, Bhagavatula, Zellers, Peters,
  Sabharwal, and Choi}]{pmlr-v119-bras20a}
Ronan~Le Bras, Swabha Swayamdipta, Chandra Bhagavatula, Rowan Zellers, Matthew
  Peters, Ashish Sabharwal, and Yejin Choi. 2020.
\newblock \href {http://proceedings.mlr.press/v119/bras20a.html} {Adversarial
  filters of dataset biases}.
\newblock In \emph{Proceedings of the 37th International Conference on Machine
  Learning}, volume 119 of \emph{Proceedings of Machine Learning Research},
  pages 1078--1088. PMLR.

\bibitem[{Chen et~al.(2020)Chen, Zheng, and Ji}]{chen-etal-2020-generating}
Hanjie Chen, Guangtao Zheng, and Yangfeng Ji. 2020.
\newblock \href {https://doi.org/10.18653/v1/2020.acl-main.494} {Generating
  hierarchical explanations on text classification via feature interaction
  detection}.
\newblock In \emph{Proceedings of the 58th Annual Meeting of the Association
  for Computational Linguistics}, pages 5578--5593, Online. Association for
  Computational Linguistics.

\bibitem[{Clark et~al.(2019{\natexlab{a}})Clark, Yatskar, and
  Zettlemoyer}]{clark2019don}
Christopher Clark, Mark Yatskar, and Luke Zettlemoyer. 2019{\natexlab{a}}.
\newblock Don’t take the easy way out: Ensemble based methods for avoiding
  known dataset biases.
\newblock In \emph{Proceedings of the 2019 Conference on Empirical Methods in
  Natural Language Processing and the 9th International Joint Conference on
  Natural Language Processing (EMNLP-IJCNLP)}, pages 4069--4082.

\bibitem[{Clark et~al.(2020)Clark, Yatskar, and
  Zettlemoyer}]{clark2020learning}
Christopher Clark, Mark Yatskar, and Luke Zettlemoyer. 2020.
\newblock \href {http://arxiv.org/abs/2011.03856} {Learning to model and ignore
  dataset bias with mixed capacity ensembles}.

\bibitem[{Clark et~al.(2019{\natexlab{b}})Clark, Khandelwal, Levy, and
  Manning}]{clark-etal-2019-bert}
Kevin Clark, Urvashi Khandelwal, Omer Levy, and Christopher~D. Manning.
  2019{\natexlab{b}}.
\newblock \href {https://doi.org/10.18653/v1/W19-4828} {What does {BERT} look
  at? an analysis of {BERT}{'}s attention}.
\newblock In \emph{Proceedings of the 2019 ACL Workshop BlackboxNLP: Analyzing
  and Interpreting Neural Networks for NLP}, pages 276--286, Florence, Italy.
  Association for Computational Linguistics.

\bibitem[{De-Arteaga et~al.(2019)De-Arteaga, Romanov, Wallach, Chayes, Borgs,
  Chouldechova, Geyik, Kenthapadi, and Kalai}]{De_Arteaga_2019}
Maria De-Arteaga, Alexey Romanov, Hanna Wallach, Jennifer Chayes, Christian
  Borgs, Alexandra Chouldechova, Sahin Geyik, Krishnaram Kenthapadi, and
  Adam~Tauman Kalai. 2019.
\newblock \href {https://doi.org/10.1145/3287560.3287572} {Bias in bios}.
\newblock \emph{Proceedings of the Conference on Fairness, Accountability, and
  Transparency}.

\bibitem[{Devlin et~al.(2019)Devlin, Chang, Lee, and
  Toutanova}]{devlin-etal-2019-bert}
Jacob Devlin, Ming-Wei Chang, Kenton Lee, and Kristina Toutanova. 2019.
\newblock \href {https://doi.org/10.18653/v1/N19-1423} {{BERT}: Pre-training of
  deep bidirectional transformers for language understanding}.
\newblock In \emph{Proceedings of the 2019 Conference of the North {A}merican
  Chapter of the Association for Computational Linguistics: Human Language
  Technologies, Volume 1 (Long and Short Papers)}, pages 4171--4186,
  Minneapolis, Minnesota. Association for Computational Linguistics.

\bibitem[{Du et~al.(2021)Du, Manjunatha, Jain, Deshpande, Dernoncourt, Gu, Sun,
  and Hu}]{DBLP:journals/corr/abs-2103-06922}
Mengnan Du, Varun Manjunatha, Rajiv Jain, Ruchi Deshpande, Franck Dernoncourt,
  Jiuxiang Gu, Tong Sun, and Xia Hu. 2021.
\newblock \href {https://arxiv.org/abs/2103.06922} {Towards interpreting and
  mitigating shortcut learning behavior of {NLU} models}.
\newblock In \emph{NAACL 2021}.

\bibitem[{Geirhos et~al.(2020)Geirhos, Jacobsen, Michaelis, Zemel, Brendel,
  Bethge, and Wichmann}]{Geirhos_2020}
Robert Geirhos, Jörn-Henrik Jacobsen, Claudio Michaelis, Richard Zemel,
  Wieland Brendel, Matthias Bethge, and Felix~A. Wichmann. 2020.
\newblock \href {https://doi.org/10.1038/s42256-020-00257-z} {Shortcut learning
  in deep neural networks}.
\newblock \emph{Nature Machine Intelligence}, 2(11):665–673.

\bibitem[{Hao et~al.(2020)Hao, Dong, Wei, and Xu}]{hao2020self-attention}
Yaru Hao, Li~Dong, Furu Wei, and Ke~Xu. 2020.
\newblock \href
  {https://www.microsoft.com/en-us/research/publication/self-attention-attribution-interpreting-information-interactions-inside-transformer/}
  {Self-attention attribution: Interpreting information interactions inside
  transformer}.
\newblock In \emph{AAAI 2021}.

\bibitem[{He et~al.(2019)He, Zha, and Wang}]{he2019unlearn}
He~He, Sheng Zha, and Haohan Wang. 2019.
\newblock Unlearn dataset bias in natural language inference by fitting the
  residual.
\newblock In \emph{Proceedings of the 2nd Workshop on Deep Learning Approaches
  for Low-Resource NLP (DeepLo 2019)}, pages 132--142.

\bibitem[{He and McAuley(2016)}]{amazon_data}
Ruining He and Julian McAuley. 2016.
\newblock \href {https://doi.org/10.1145/2872427.2883037} {Ups and downs:
  Modeling the visual evolution of fashion trends with one-class collaborative
  filtering}.
\newblock In \emph{Proceedings of the 25th International Conference on World
  Wide Web}, WWW ’16.

\bibitem[{Hendrycks et~al.(2020)Hendrycks, Liu, Wallace, Dziedzic, Krishnan,
  and Song}]{ood_robustness}
Dan Hendrycks, Xiaoyuan Liu, Eric Wallace, Adam Dziedzic, Rishabh Krishnan, and
  Dawn Song. 2020.
\newblock Pretrained transformers improve out-of-distribution robustness.
\newblock In \emph{Proceedings of the 58th Annual Meeting of the Association
  for Computational Linguistics}.

\bibitem[{Jacovi et~al.(2021)Jacovi, Swayamdipta, Ravfogel, Elazar, Choi, and
  Goldberg}]{jacovi2021contrastive}
Alon Jacovi, Swabha Swayamdipta, Shauli Ravfogel, Yanai Elazar, Yejin Choi, and
  Yoav Goldberg. 2021.
\newblock \href {http://arxiv.org/abs/2103.01378} {Contrastive explanations for
  model interpretability}.

\bibitem[{Jia and Liang(2017)}]{jia-liang-2017-adversarial}
Robin Jia and Percy Liang. 2017.
\newblock \href {https://doi.org/10.18653/v1/D17-1215} {Adversarial examples
  for evaluating reading comprehension systems}.
\newblock In \emph{Proceedings of the 2017 Conference on Empirical Methods in
  Natural Language Processing}, pages 2021--2031, Copenhagen, Denmark.
  Association for Computational Linguistics.

\bibitem[{Jiaao et~al.(2021)Jiaao, Dinghan, Weizhu, and Diyi}]{hiddencut}
Chen Jiaao, Shen Dinghan, Chen Weizhu, and Yang Diyi. 2021.
\newblock Hiddencut: Simple data augmentation for natural language
  understanding with better generalizability.
\newblock In \emph{Proceedings of the 59th Annual Meeting of the Association of
  Computational Linguistics}. Association for Computational Linguistics.

\bibitem[{Jin et~al.(2020)Jin, Jin, Zhou, and Szolovits}]{textfooler}
Di~Jin, Zhijing Jin, Joey Zhou, and Peter Szolovits. 2020.
\newblock Is {BERT} really robust? {N}atural language attack on text
  classification and entailment.
\newblock In \emph{AAAI}.

\bibitem[{Khani and Liang(2021)}]{10.1145/3442188.3445883}
Fereshte Khani and Percy Liang. 2021.
\newblock \href {https://doi.org/10.1145/3442188.3445883} {Removing spurious
  features can hurt accuracy and affect groups disproportionately}.
\newblock In \emph{Proceedings of the 2021 ACM Conference on Fairness,
  Accountability, and Transparency}, FAccT '21, page 196–205, New York, NY,
  USA. Association for Computing Machinery.

\bibitem[{Kovaleva et~al.(2019)Kovaleva, Romanov, Rogers, and
  Rumshisky}]{kovaleva-etal-2019-revealing}
Olga Kovaleva, Alexey Romanov, Anna Rogers, and Anna Rumshisky. 2019.
\newblock \href {https://doi.org/10.18653/v1/D19-1445} {Revealing the dark
  secrets of {BERT}}.
\newblock In \emph{Proceedings of the 2019 Conference on Empirical Methods in
  Natural Language Processing and the 9th International Joint Conference on
  Natural Language Processing (EMNLP-IJCNLP)}, pages 4365--4374, Hong Kong,
  China. Association for Computational Linguistics.

\bibitem[{McCoy et~al.(2019)McCoy, Pavlick, and Linzen}]{mccoy-etal-2019-right}
Tom McCoy, Ellie Pavlick, and Tal Linzen. 2019.
\newblock \href {https://doi.org/10.18653/v1/P19-1334} {Right for the wrong
  reasons: Diagnosing syntactic heuristics in natural language inference}.
\newblock In \emph{Proceedings of the 57th Annual Meeting of the Association
  for Computational Linguistics}, pages 3428--3448, Florence, Italy.
  Association for Computational Linguistics.

\bibitem[{Miller(1995)}]{wordnet}
George~A. Miller. 1995.
\newblock \href {https://doi.org/10.1145/219717.219748} {Wordnet: A lexical
  database for english}.
\newblock \emph{Commun. ACM}, 38(11):39–41.

\bibitem[{Minderer et~al.(2020)Minderer, Bachem, Houlsby, and
  Tschannen}]{minderer2020automatic}
Matthias Minderer, Olivier Bachem, Neil Houlsby, and Michael Tschannen. 2020.
\newblock Automatic shortcut removal for self-supervised representation
  learning.
\newblock In \emph{Proceedings of the 37th International Conference on Machine
  Learning}.

\bibitem[{Mrk\v{s}i\'c et~al.(2016)Mrk\v{s}i\'c, {\'O S\'eaghdha}, Thomson,
  Ga\v{s}i\'c, Rojas-Barahona, Su, Vandyke, Wen, and Young}]{counter-fitting}
Nikola Mrk\v{s}i\'c, Diarmuid {\'O S\'eaghdha}, Blaise Thomson, Milica
  Ga\v{s}i\'c, Lina Rojas-Barahona, Pei-Hao Su, David Vandyke, Tsung-Hsien Wen,
  and Steve Young. 2016.
\newblock Counter-fitting word vectors to linguistic constraints.
\newblock In \emph{Proceedings of HLT-NAACL}.

\bibitem[{Narang et~al.(2020)Narang, Raffel, Lee, Roberts, Fiedel, and
  Malkan}]{DBLP:journals/corr/abs-2004-14546}
Sharan Narang, Colin Raffel, Katherine Lee, Adam Roberts, Noah Fiedel, and
  Karishma Malkan. 2020.
\newblock \href {https://arxiv.org/abs/2004.14546} {Wt5?! training text-to-text
  models to explain their predictions.}
\newblock \emph{CoRR}, abs/2004.14546.

\bibitem[{Niu et~al.(2020)Niu, Mathur, Dinu, and
  Al-Onaizan}]{niu-etal-2020-evaluating}
Xing Niu, Prashant Mathur, Georgiana Dinu, and Yaser Al-Onaizan. 2020.
\newblock \href {https://doi.org/10.18653/v1/2020.acl-main.755} {Evaluating
  robustness to input perturbations for neural machine translation}.
\newblock In \emph{Proceedings of the 58th Annual Meeting of the Association
  for Computational Linguistics}, pages 8538--8544, Online. Association for
  Computational Linguistics.

\bibitem[{Pan and Yang(2010)}]{pan_transfer_learning_survey}
Sinno~Jialin Pan and Qiang Yang. 2010.
\newblock \href {https://doi.org/10.1109/TKDE.2009.191} {A survey on transfer
  learning}.
\newblock \emph{IEEE Transactions on Knowledge and Data Engineering},
  22(10):1345--1359.

\bibitem[{Pruthi et~al.(2020)Pruthi, Gupta, Dhingra, Neubig, and
  Lipton}]{pruthi-etal-2020-learning}
Danish Pruthi, Mansi Gupta, Bhuwan Dhingra, Graham Neubig, and Zachary~C.
  Lipton. 2020.
\newblock \href {https://doi.org/10.18653/v1/2020.acl-main.432} {Learning to
  deceive with attention-based explanations}.
\newblock In \emph{Proceedings of the 58th Annual Meeting of the Association
  for Computational Linguistics}, pages 4782--4793, Online. Association for
  Computational Linguistics.

\bibitem[{Ribeiro et~al.(2016)Ribeiro, Singh, and Guestrin}]{LIME}
Marco~Tulio Ribeiro, Sameer Singh, and Carlos Guestrin. 2016.
\newblock "why should {I} trust you?": Explaining the predictions of any
  classifier.
\newblock In \emph{Proceedings of the 22nd {ACM} {SIGKDD} International
  Conference on Knowledge Discovery and Data Mining, San Francisco, CA, USA,
  August 13-17, 2016}, pages 1135--1144.

\bibitem[{Ribeiro et~al.(2020)Ribeiro, Wu, Guestrin, and
  Singh}]{ribeiro-etal-2020-beyond}
Marco~Tulio Ribeiro, Tongshuang Wu, Carlos Guestrin, and Sameer Singh. 2020.
\newblock \href {https://doi.org/10.18653/v1/2020.acl-main.442} {Beyond
  accuracy: Behavioral testing of {NLP} models with {C}heck{L}ist}.
\newblock In \emph{Proceedings of the 58th Annual Meeting of the Association
  for Computational Linguistics}, pages 4902--4912, Online. Association for
  Computational Linguistics.

\bibitem[{Sagawa et~al.(2020)Sagawa, Raghunathan, Koh, and
  Liang}]{sagawa2020investigation}
Shiori Sagawa, Aditi Raghunathan, Pang~Wei Koh, and Percy Liang. 2020.
\newblock \href {http://arxiv.org/abs/2005.04345} {An investigation of why
  overparameterization exacerbates spurious correlations}.

\bibitem[{Socher et~al.(2013)Socher, Perelygin, Wu, Chuang, Manning, Ng, and
  Potts}]{socher-etal-2013-recursive}
Richard Socher, Alex Perelygin, Jean Wu, Jason Chuang, Christopher~D. Manning,
  Andrew Ng, and Christopher Potts. 2013.
\newblock \href {https://www.aclweb.org/anthology/D13-1170} {Recursive deep
  models for semantic compositionality over a sentiment treebank}.
\newblock In \emph{Proceedings of the 2013 Conference on Empirical Methods in
  Natural Language Processing}, pages 1631--1642, Seattle, Washington, USA.
  Association for Computational Linguistics.

\bibitem[{Srivastava et~al.(2020)Srivastava, Hashimoto, and
  Liang}]{pmlr-v119-srivastava20a}
Megha Srivastava, Tatsunori Hashimoto, and Percy Liang. 2020.
\newblock \href {http://proceedings.mlr.press/v119/srivastava20a.html}
  {Robustness to spurious correlations via human annotations}.
\newblock In \emph{Proceedings of the 37th International Conference on Machine
  Learning}, volume 119 of \emph{Proceedings of Machine Learning Research},
  pages 9109--9119. PMLR.

\bibitem[{Sundararajan et~al.(2017)Sundararajan, Taly, and Yan}]{pig}
Mukund Sundararajan, Ankur Taly, and Qiqi Yan. 2017.
\newblock \href {http://proceedings.mlr.press/v70/sundararajan17a.html}
  {Axiomatic attribution for deep networks}.
\newblock In \emph{Proceedings of the 34th International Conference on Machine
  Learning}, volume~70 of \emph{Proceedings of Machine Learning Research},
  pages 3319--3328. PMLR.

\bibitem[{Tsipras et~al.(2019)Tsipras, Santurkar, Engstrom, Turner, and
  Madry}]{tsipras2018robustness}
Dimitris Tsipras, Shibani Santurkar, Logan Engstrom, Alexander Turner, and
  Aleksander Madry. 2019.
\newblock \href {https://openreview.net/forum?id=SyxAb30cY7} {Robustness may be
  at odds with accuracy}.
\newblock In \emph{International Conference on Learning Representations}.

\bibitem[{Tu et~al.(2020)Tu, Lalwani, Gella, and He}]{tu20tacl}
Lifu Tu, Garima Lalwani, Spandana Gella, and He~He. 2020.
\newblock \href {https://arxiv.org/abs/2007.06778} {An empirical study on
  robustness to spurious correlations using pre-trained language models}.
\newblock \emph{Transactions of the Association of Computational Linguistics}.

\bibitem[{Wang et~al.(2019)Wang, Gan, Liu, Liu, Gao, and
  Wang}]{wang-etal-2019-adversarial-domain}
Huazheng Wang, Zhe Gan, Xiaodong Liu, Jingjing Liu, Jianfeng Gao, and Hongning
  Wang. 2019.
\newblock \href {https://doi.org/10.18653/v1/D19-1254} {Adversarial domain
  adaptation for machine reading comprehension}.
\newblock In \emph{Proceedings of the 2019 Conference on Empirical Methods in
  Natural Language Processing and the 9th International Joint Conference on
  Natural Language Processing (EMNLP-IJCNLP)}.

\bibitem[{Wang and Culotta(2020{\natexlab{a}})}]{wang-culotta-2020-identifying}
Zhao Wang and Aron Culotta. 2020{\natexlab{a}}.
\newblock \href {https://doi.org/10.18653/v1/2020.findings-emnlp.308}
  {Identifying spurious correlations for robust text classification}.
\newblock In \emph{Findings of the Association for Computational Linguistics:
  EMNLP 2020}, pages 3431--3440, Online. Association for Computational
  Linguistics.

\bibitem[{Wang and Culotta(2020{\natexlab{b}})}]{zhao_spurious_2020}
Zhao Wang and Aron Culotta. 2020{\natexlab{b}}.
\newblock \href {https://arxiv.org/abs/2012.10040} {Robustness to spurious
  correlations in text classification via automatically generated
  counterfactuals}.
\newblock In \emph{AAAI}.

\bibitem[{Zhou and Bansal(2020)}]{zhou-bansal-2020-towards}
Xiang Zhou and Mohit Bansal. 2020.
\newblock \href {https://doi.org/10.18653/v1/2020.acl-main.773} {Towards
  robustifying {NLI} models against lexical dataset biases}.
\newblock In \emph{Proceedings of the 58th Annual Meeting of the Association
  for Computational Linguistics}, pages 8759--8771, Online. Association for
  Computational Linguistics.

\end{thebibliography}
